\documentclass{iise}

\usepackage{multirow}

 \conference{Proceedings of the IISE Annual Conference \& Expo 2023 \\
 K. Babski-Reeves, B. Eksioglu, D. Hampton, eds.}

\title{\titlesize Eye Disease Classification Using Deep Learning Techniques}

\author{
Tareq Babaqi, Manar Jaradat, Ayse Erdem Yildirim, Saif H. Al-Nimer, and Daehan Won
\vspace{5mm}\\Department of Systems Science \& Industrial Engineering\\State University of New York at Binghamton,Binghamton, NY, 13902 \\}

\authorlist{Babaqi.T, Jaradat. M, Yildirim. A, Al-Nimer. S, and Won. D}
\abstractID{1944}

\begin{document}
\maketitle

\begin{abstract}
{\small Eye is the essential sense organ for vision function. Due to the fact that certain eye disorders might result in vision loss, it is essential to diagnose and treat eye diseases early on. By identifying common eye illnesses and performing an eye check, eye care providers can safeguard patients against vision loss or blindness. Convolutional neural networks (CNN) and transfer learning were employed in this study to discriminate between a normal eye and one with diabetic retinopathy, cataract, or glaucoma disease. Using transfer learning for multi-class classification, high accuracy was achieved at 94\% while the traditional CNN achieved 84\% rate.}
\end{abstract}

\section*{Keywords}
Eye Disease, Image classification, Data Mining, CNN, Transfer learning.

\section{Introduction}
Patients with vision problems suffer from significantly reduced quality of life. Diabetes retinopathy and glaucoma are examples of eye diseases. Globally, 64.3 million people had glaucoma in 2013 \cite{ref1_tham2014global}. Early detection of eye diseases can aid in proper treatment or, at the very least, stop these conditions from worsening. However, it is restricted due to the lack of awareness, shortage of ophthalmologists, and high consultation costs., Hence, automated screening is critical \cite{ref2_chelaramani2021multi}. This paper will utilize machine learning techniques to distinguish a normal eye from one with a disease. 

Using eye pictures, numerous research has attempted to anticipate and/or categorize a healthy eye from one with a disease. \cite{ref8_s21165283} presented a robust automated method based on a proprietary CenterNet model and a DenseNet-100 feature extractor for detecting and classifying diabetic retinopathy and diabetic macular edema lesions. They attained accuracies of 97.93 \% and 98.10 \%, respectively, while evaluating their methodology using the APTOS-2019 and IDRiD benchmark datasets. \cite{ref3_smaida2019comparative} provided three classification techniques for multiple classes: Convolutional Neural Network (CNN), Visual Geometry Group 16 (VGG16), and Inception V3. Using a confusion matrix, the precision of each method is measured and compared.

\cite{ref5_grassmann2018deep} developed an algorithm for categorizing age-related macular degeneration using a substantial collection of color fundus pictures as their data source. The validation is carried out utilizing a population-based study that is cross-sectional. Within the Age-Related Eye Disease Study context, 120,656 color fundus pictures were manually graded from the eyes of 3654 participants in Age-Related Eye Disease Studies (AREDS). In their 2016 paper, \cite{ref6_bhatia2016diagnosis} discussed many strategies that can be used to construct an automated system that can detect cases of diabetic retinopathy in people who have diabetes. Its purpose was to provide ophthalmologists with a tool to aid in the early detection of diabetic retinopathy symptoms. In addition, various technologies employed for the diagnosis and early detection of diabetic eye illness are discussed.

Diabetic eye disease (DED) categorization relies heavily on image processing; hence \cite{ref7_sarki2021image} offered a comprehensive study on the topic. Picture quality enrichment, image segmentation, image augmentation (geometric transformation), and classification comprise the suggested automated classification framework for DED. The best results were achieved by combining conventional approaches to image processing with a freshly developed convolution neural network (CNN) structure. When applied to DED classification problems, the newly constructed CNN combined with the conventional image processing method showed the best performance with accuracy. The experimental data indicated satisfactory levels of precision, sensitivity, and specificity.

\section{Problem and Data Description}
Eye disease early prediction has many restrictions and limitations, such as a shortage of experts and high consultation costs. As a result, machine learning should be used to optimize the present system in order to give maximum comfort to the patient/optometrist and enhance health care. This paper will use transfer learning, “a machine learning technique where a model developed for a task is reused as the starting point for a model on a second task” \cite{ref4_brownlee_2019}, to classify eye disease. Since no previous papers have discussed using transfer learning in eye image classification, this can be considered a good contribution to this work.

\begin{figure}[ht]
\centering
\includegraphics[scale=0.5]{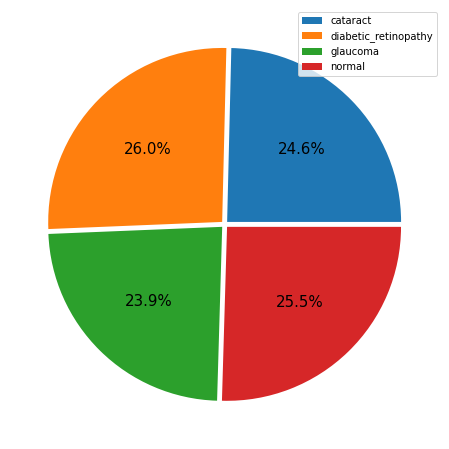}
\caption{Image distribution between classes}
\label{fig:image_distribution}
\end{figure}

Data used in this paper were collected from the following resources \cite{ref9_data3030025}, \cite{ref10_fundus_image}, \cite{ref11_ocular_image} and \cite{ref12_retinal_image}. The dataset includes around 4200 colored images for normal eyes, for eyes with cataracts, diabetic retinopathy, and glaucoma. Figure \ref{fig:image_distribution} depicts the percentage of images that belong to each class. Images that belong to each category were stored in a separate folder and labeled accordingly. Images were rescaled from $512 \times 512 \times 3$ to $224 \times 224 \times 3$ to help improve the model's generalization performance. By rescaling the images to a smaller size, we can reduce the complexity of the model and prevent overfitting, which can improve the model's ability to generalize to new images. Then the data was split into 3 subsets; 70 \% training, 20 \% testing, and 10\% validation.

\section{Models and Approaches} \label{s:numerical}
\subsection{\emph{Convolutional Neural Network}} \label{ss: cnn}
LeCun first introduces the convolutional neural network (CNN) to detect handwritten numbers \cite{ref20_lecun1998gradient}. Instated of simple matrix multiplication, the authors use convolution operation in the simple architecture of CNN. Nevertheless, the architecture of CNN could be enhanced by adding some essential layers, such as convolutional layer (Conv), rectified Linear unit (ReLU), batch normalization, pooling layers (Pool), fully connected layers (FC) \cite{ref21_gour2021multi}. Thus, it would be extremely useful when dealing with complex problems that require a tremendous amount of data, such as image segmentation, medical image processing, image enhancement, pattern recognition, classification, etc. The performance of a CNN can be improved by having access to more data during training, as it allows the model to learn a wider variety of features and patterns in the data. However, training CNNs on smaller datasets and achieving reasonable performance is also possible, especially if the dataset is carefully curated to be diverse and representative of the target domain. Several CNN architectures have been developed in the literature to deal with images, including the databases that may be used in computer vision tasks. Several CNN architectures are developed in the literature, e.g., LeNet, AlexNet, VGGNet, GoogleNet, and ResNet, to deal with the images \cite{ref23_tan2019efficientnet}. Each of these architectures has its own unique characteristics and strengths, and they have been shown to be effective for different types of image classification tasks.

\begin{figure}[h!]
\centering
\includegraphics[scale=1.5]{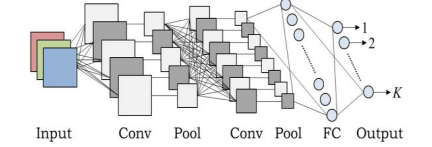}
\caption{General CNN architecture}
\label{fig:cnn_model}
\end{figure}

Figure \ref{fig:cnn_model} demonstrates a CNN architecture of the aforementioned layers, briefly explained below.
\begin{itemize}
\item Convolutional layer: This layer applies filters via the convolution operator to input data in order to find important features. Later, these features are passed to the next layer.
\item Rectified Linear unit (ReLU): This layer uses the non-saturating activation function in order to replace negative values in convolutional layers with zero. Thus, the training process is expedited and improved.
\item Pooling layers: These layers reduce the dimensions of an image. It creates a window for neurons and chooses each window's max value. 
\item Fully connected layers: This is the last layer and fully connects each neuron in one layer to every neuron in another layer. 
\end{itemize}

One of the advantages of CNN is that it eliminates the manual feature extraction process. Unlike machine learning algorithms which require a separate feature extraction process, CNN does feature extraction automatically. The feature extraction is done in the convolutional, ReLU, and pooling layers. Finally, the classification is completed in the fully connected layer.

\subsection{\emph{Transfer Learning}} \label{ss: transfer}
As stated in section \ref{ss: cnn}, CNN is complex and requires a huge dataset to learn. Training CNN models with such datasets from scratch requires enormous computational power and resources. Therefore, regardless of the domain, CNN model parameters and weights are transferred from existing network models. This is done in order to shorten the training process and improve model performance. Transferring this knowledge from another network is called transfer learning \cite{ref22_krishna2019deep}. Transfer learning can be applied in two ways. First, Fine-tuning: In fine-tuning, the pre-trained model is adapted to the new task by retraining some or all of the model's layers on the new dataset. Typically, the earlier layers of the pre-trained model are frozen, while the later layers are retrained to learn task-specific features. Second, Feature extraction: In feature extraction, the pre-trained model is used as a fixed feature extractor, and the output of one or more layers of the model is used as input to a new model trained on the new dataset. In both cases, transfer learning can be beneficial because the pre-trained model has already learned a set of feature representations that can be useful for the new task. This can save time and computational resources compared to training a new model from scratch and can also improve the generalization performance of the new model by leveraging the pre-trained model's ability to learn useful features. This paper used the first way as the task is so related to the original task of the pre-trained model, while the second can be used when the task is different from the original task for which the pre-trained model was trained. To use a pre-trained model, select the source, reuse, and tune the model.

\subsection{\emph{Used CNN model}} \label{ss: used_model}
In this investigation, the CNN model is developed by transfer learning using a pre-trained EfficientNet CNN architecture model. EfficientNet is a CNN architecture and scaling technique that uniformly scales all depth, breadth, and resolution parameters \cite{ko}. It does this by using a compound coefficient to scale the dimensions. In contrast to the traditional technique, which scales these variables arbitrarily, the EfficientNet scaling method scales the network breadth, depth, and resolution uniformly by using a set of predetermined scaling factors.

\section{Results}\label{s:results}
In this section, the process for establishing tests to evaluate the methodologies and documenting the results gained is discussed in detail. The results of the tests that have been presented should help answer the following question: Does Transfer Learning contributes to the improvement of CNN's accuracy?

The evaluation metrics used in this research are Precision, recall, and F1-score, as they are commonly used metrics in machine learning and information retrieval to evaluate the performance of a classification model. Precision is the fraction of true positives (correctly identified instances of a particular class) out of all positive predictions made by the model. In other words, precision measures how many items predicted to be positive are positive. Precision is calculated as: TP / (TP + FP), where TP is the number of true positives and FP is the number of false positives. Recall is the fraction of true positives out of all instances that belong to a particular class. In other words, recall measures how many of the positive instances in the dataset were correctly identified by the model. Recall is calculated as: TP / (TP + FN). The F1-score is the harmonic mean of precision and recall. It is a way to combine precision and recall into a single metric that gives equal weight to both measures. The F1-score is calculated as: 2 * (precision * recall) / (precision + recall). Also, Accuracy is another commonly used metric in machine learning and information retrieval. It measures the overall performance of a classification model by calculating the fraction of correctly classified instances (both true positives and true negatives) out of the total number of instances in the dataset. Accuracy is calculated as: (TP + TN) / (TP + TN + FP + FN).

Table \ref{tab1} shows that the transfer learning model, which used the pre-trained model, achieved higher accuracy of 94 \% compared to the CNN model, which achieved a classification accuracy of 84 \%. It can also be seen that the F1-score of Diabetic retinopathy is 100 \%, which shows no misclassification exists in this disease class by the CNN model, while 99 \% in the Transfer learning model, which means only one misclassification exists. The results of the other classes (Normal, Cataract, and Glaucoma) are better for the Transfer Learning model in all evaluation metrics in Table \ref{tab1}.

 \begin{table}[htb]
\caption{Evaluation metrics for the used classification algorithms}\label{tab1}
\vspace{-0.7cm}
\begin{center}
\begin{tabular}{lllcccc}
\hline
\multicolumn{1}{}{l}Model  & Class& Precision (\%) & Recall (\%) & F1-score (\%) & Accuracy (\%)\\ \hline
\multirow{3}{*}{CNN} & cataract  & 87	& 96 & 91 & \multirow{3}{*}{84}\\ \
 & Diabetic retinopathy & 100 & 100	& 100 \\ \
 & Glaucoma & 67	& 85 & 75\\ 
 & Normal & 86 & 56	& 68 \\ \cline{1-6}
 \multirow{3}{*}{Transfer Learning} & cataract & 97 & 96	& 96 & \multirow{3}{*}{94}\\ \
 & Diabetic retinopathy & 100 & 99 & 99 \\ \
 & Glaucoma & 94	& 85 & 89 \\ 
 & Normal & 86 & 96 & 91 \\ \cline{1-6}
\end{tabular}
\end{center}
\end{table}

\begin{figure}[htb]
\centering
\includegraphics[scale=.45]{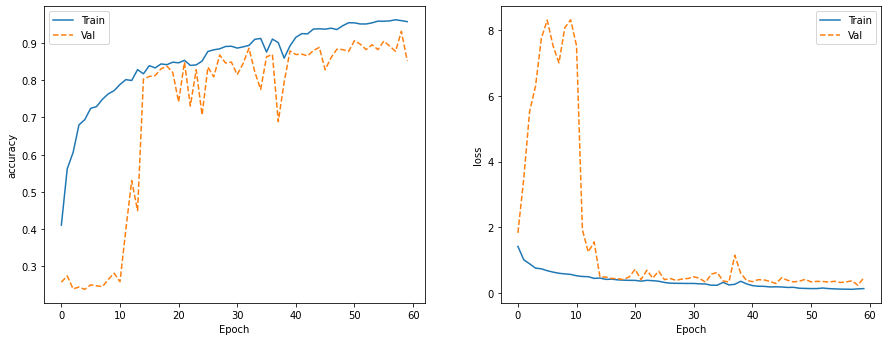}
\caption{Training accuracy and loss for CNN}
\label{fig:cnn_loss}
\end{figure}

\begin{figure}[htb]
\centering
\includegraphics[scale=.45]{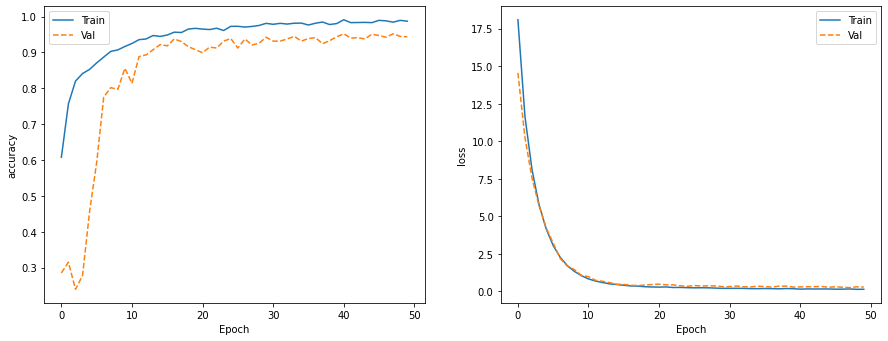}
\caption{Training accuracy and loss for transfer learning}
\label{fig:trsfr_loss}
\end{figure}

Figure \ref{fig:cnn_loss} shows how the accuracy and loss improved when the CNN model's epochs increased. Figure \ref{fig:trsfr_loss} shows the improvement of the accuracy and loss in the Transfer Learning model, which shows a smoother improvement than CNN (with a lot of fluctuation). 

\begin{figure}[htb]
         \centering
         \includegraphics[width=\textwidth]{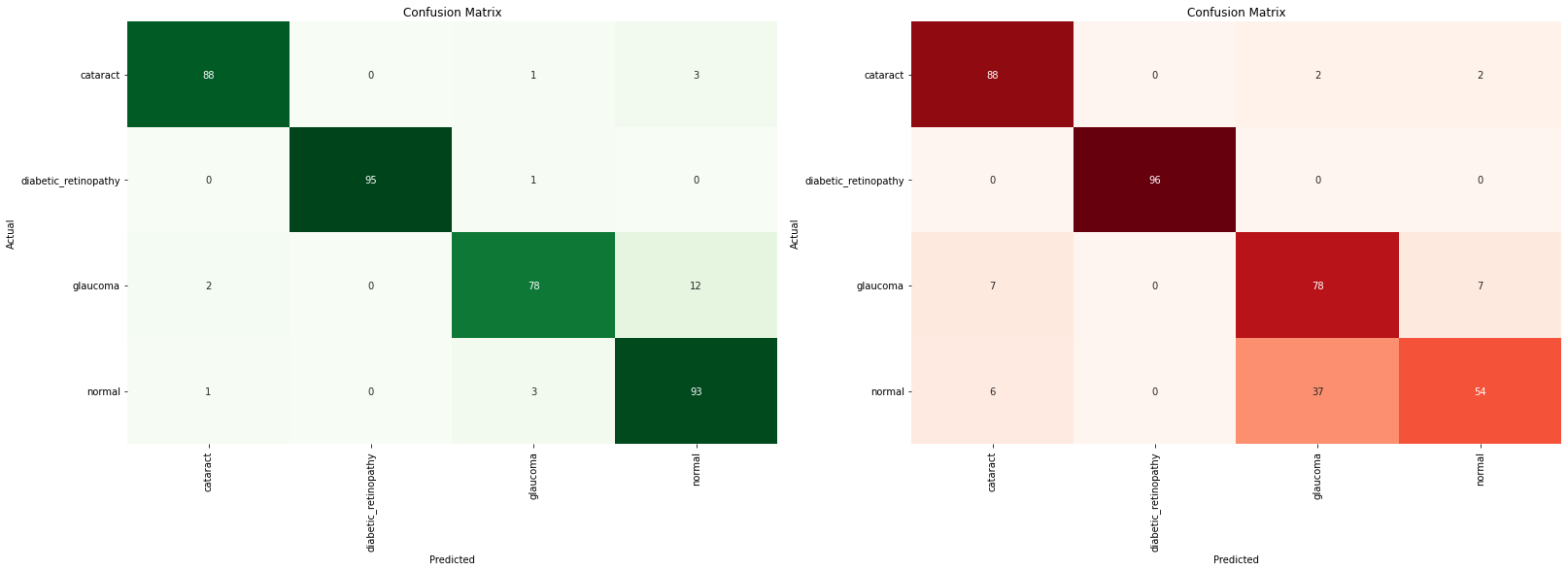}

     \hfill
     \caption{Confusion matrices for both Transfer learning and CNN model}
        \label{fig:image.png}
\end{figure}

Both models' misclassification data points and correct classification points are summarized in confusion matrices in figure \ref{fig:image.png}. The left is the confusion matrix of the transfer learning model, while the right is the CNN model's confusion matrix. This evidently shows the benefits and the power of Transfer Learning.

\section{Conclusion}\label{s:conclusion}
The eye is one of the most important sense organs in human anatomy, and losing vision would significantly affect the quality of life. Besides, the eye could also indicate some serious health issues. Unfortunately, many people may not be aware of these health problems due to the lack of ophthalmologists and access to eye care. Nevertheless, the advent of deep learning and image processing could immensely help in detecting and diagnosing these diseases. This study used CNN and transfer learning to detect three eye diseases: cataracts, diabetic retinopathy, and glaucoma. While CNN alone is not good enough for classifying eye diseases, its performance significantly increased with transfer learning. The pre-trained model of EfficientNet is transferred to the CNN model using transfer learning. 

For future research, we plan to investigate the efficacy of transfer learning on a more diverse dataset that includes other types of eye diseases. And explore other applications of transfer learning in disease detection tasks.



\begin{thebibliography}{1}


\bibitem{ref1_tham2014global}Tham, Y., Li, X., Wong, T., Quigley, H., Aung, T. \& Cheng, C. Global prevalence of glaucoma and projections of glaucoma burden through 2040: a systematic review and meta-analysis. {\em Ophthalmology}. \textbf{121}, 2081-2090 (2014)

\bibitem{ref2_chelaramani2021multi}Chelaramani, S., Gupta, M., Agarwal, V., Gupta, P. \& Habash, R. Multi-task knowledge distillation for eye disease prediction. {\em Proceedings Of The IEEE/CVF Winter Conference On Applications Of Computer Vision}. pp. 3983-3993 (2021)

\bibitem{ref8_s21165283}Nazir, T., Nawaz, M., Rashid, J., Mahum, R., Masood, M., Mehmood, A., Ali, F., Kim, J., Kwon, H. \& Hussain, A. Detection of Diabetic Eye Disease from Retinal Images Using a Deep Learning Based CenterNet Model. {\em Sensors}. \textbf{21} (2021), https://www.mdpi.com/1424-8220/21/16/5283

\bibitem{ref3_smaida2019comparative}Smaida, M. \& Serhii, Y. Comparative Study of Image Classification Algorithms for Eyes Diseases Diagnostic. {\em International Journal Of Innovative Science And Research Technology}. \textbf{4} (2019)

\bibitem{ref5_grassmann2018deep}Grassmann, F., Mengelkamp, J., Brandl, C., Harsch, S., Zimmermann, M., Linkohr, B., Peters, A., Heid, I., Palm, C. \& Weber, B. A deep learning algorithm for prediction of age-related eye disease study severity scale for age-related macular degeneration from color fundus photography. {\em Ophthalmology}. \textbf{125}, 1410-1420 (2018)

\bibitem{ref6_bhatia2016diagnosis}Bhatia, K., Arora, S. \& Tomar, R. Diagnosis of diabetic retinopathy using machine learning classification algorithm. {\em 2016 2nd International Conference On Next Generation Computing Technologies (NGCT)}. pp. 347-351 (2016)

\bibitem{ref7_sarki2021image}Sarki, R., Ahmed, K., Wang, H., Zhang, Y., Ma, J. \& Wang, K. Image preprocessing in classification and identification of diabetic eye diseases. {\em Data Science And Engineering}. \textbf{6}, 455-471 (2021)

\bibitem{ref4_brownlee_2019}Brownlee, J. A gentle introduction to transfer learning for Deep learning. {\em MachineLearningMastery.com}. (2019,9), https://machinelearningmastery.com/transfer-learning-for-deep-learning/


\bibitem{ref9_data3030025}Porwal, P., Pachade, S., Kamble, R., Kokare, M., Deshmukh, G., Sahasrabuddhe, V. \& Meriaudeau, F. Indian Diabetic Retinopathy Image Dataset (IDRiD): A Database for Diabetic Retinopathy Screening Research. {\em Data}. \textbf{3} (2018), https://www.mdpi.com/2306-5729/3/3/25

\bibitem{ref10_fundus_image}HRF High-Resolution Fundus Image Database. {\em Www5.cs.fau.de}. (2022,12), https://www5.cs.fau.de/research/data/fundus-images/

\bibitem{ref11_ocular_image}Larxel Ocular disease recognition. {\em Www.kaggle.com}. (2020,9), https://www.kaggle.com/datasets/andrewmvd/ocular-disease-recognition-odir5k

\bibitem{ref12_retinal_image}Larxel Retinal Disease Classification. {\em Www.kaggle.com}. (2021,8), https://www.kaggle.com/datasets/andrewmvd/retinal-disease-classification

\bibitem{ref20_lecun1998gradient}LeCun, Y., Bottou, L., Bengio, Y. \& Haffner, P. Gradient-based learning applied to document recognition. {\em Proceedings Of The IEEE}. \textbf{86}, 2278-2324 (1998)

\bibitem{ref21_gour2021multi}Gour, N. \& Khanna, P. Multi-class multi-label ophthalmological disease detection using transfer learning based convolutional neural network. {\em Biomedical Signal Processing And Control}. \textbf{66} pp. 102329 (2021)

\bibitem{ref23_tan2019efficientnet}Tan, M. \& Le, Q. Efficientnet: Rethinking model scaling for convolutional neural networks. {\em International Conference On Machine Learning}. pp. 6105-6114 (2019)

\bibitem{ref22_krishna2019deep}Krishna, S. \& Kalluri, H. Deep learning and transfer learning approaches for image classification. {\em International Journal Of Recent Technology And Engineering (IJRTE)}. \textbf{7}, 427-432 (2019)

\bibitem{ko}Koonce, B. \& Koonce, B. EfficientNet. {\em Convolutional Neural Networks With Swift For Tensorflow: Image Recognition And Dataset Categorization}. pp. 109-123 (2021)

 
\end{thebibliography}
\end{document}